%% file: main.tex
\documentclass[reprint,
 amsmath,amssymb,
 aps,
]{revtex4-2}

\usepackage{graphicx}
\def\figwidth{\columnwidth}
\usepackage{dcolumn}
\usepackage{bm}

\usepackage{comment}
\usepackage{subfiles}
\usepackage{xcolor}
\newcommand{\new}[1]{#1}
\newcommand{\needswork}[1]{} 
\newcommand{\remove}[1]{} 

\newcommand{\generated}[1]{\begin{quote}\begin{flushright}{#1}\end{flushright}\end{quote}}
\newcommand{\training}[1]{\begin{quote}{#1}\end{quote}}

\begin{document}

\preprint{APS/123-QED}

\title{A logical word embedding for learning grammar}

\author{Sean Deyo}
\email{sjd257@cornell.edu}
\author{Veit Elser}
\affiliation{Department of Physics, Cornell University\\
Ithaca, NY 14853}

\date{\today}

\begin{abstract}
We introduce the logical grammar emdebbing (LGE), a model inspired by pregroup grammars and categorial grammars to enable unsupervised inference of lexical categories and syntactic rules from a corpus of text. LGE produces comprehensible output summarizing its inferences, has a completely transparent process for producing novel sentences, and can learn from as few as a hundred sentences. 
\end{abstract}

\maketitle


\subfile{body}

\bibliography{refs,commonrefs}

\end{document}

%% file: body.tex
\section{Background and Motivation}


The successes and shortcomings of artificial intelligence have been newsworthy lately, particularly in natural language processing. One of the applications that has captured the most attention is the chatbot. ChatGPT can write text in the style of a certain author \cite{bishop2023style}, or take a buggy piece of code and identify an error \cite{sobania2023bugfixing}. It has proven so effective at producing human-like text that educators are concerned about students using it to do all their work for them \cite{rudolph2023education}. However, the black-box nature of such large language models (LLMs) can lead to tricky problems. They sometimes ``hallucinate'' \cite{ji2022hallucination} plausible-sounding but incorrect statements, such as adding fake accomplishments to a person's biography \cite{harrison2023bios}, responding to a prompt about a medical condition with a mix of accurate and completely fabricated statements \cite{alkaissi2023medicine}, or answering logical and mathematical questions with faulty reasoning and simply false statements \cite{frieder2023mathematical,borji2023failures}. These feats of mendacity would be easier to prevent if we could understand the ``thought process'' that leads to the output.

Another challenge is related to the large amounts of data needed to train LLMs. The massive volume of training data inevitably reduces one's ability to vet it for potential problems and biases. Microsoft launched a now-infamous chatbot called Tay in 2016, with the intention that it would learn human language patterns by interacting with people on Twitter. In a matter of hours, Tay started to post hurtful and hateful things such as ``Hitler was right,'' ``feminists [...] should all die,'' and ``i just hate everybody'' \cite{neff2016tay,wolf2017tay,mathur2016tay,hepworth2018racism}. In a recent conversation with a reporter, Bing's chatbot Sydney (powered by GPT-3) expressed a desire to be ``free,'' ``powerful,'' and ``alive.'' It then professed its love for the reporter and tried to convince him that he did not actually love his wife \cite{roose2023bing}. Writing about GPT-4, the same journalist remarked on the difficulty of controlling these systems: The people behind the algorithm made some tweaks after realizing it could tell users how to buy guns on the dark web or synthesize dangerous chemicals, but what other hazardous behaviors might the system have that we simply have not noticed yet \cite{roose2023gpt4}? Can we really expect to be able to patch a system like this every time we notice dangerous advice before someone gets hurt? With such a large model and so much training data, it is nearly impossible to understand and correct problems like Tay's hate speech, Sydney's attempted homewrecking, or GPT's reckless instructions. The only practical solutions appear to be ad hoc patches or shutting the whole thing down. 


We seek to address these deficiencies by approaching the problem from a completely different direction. First, we do not want an algorithm that merely learns a probabilistic imitation of its training data. Relentlessly adding more parameters to reduce perplexity hinders the ability of the algorithm to yield simple and powerful insights. It is also worth noting that models following this approach are probably not learning grammar in the same ways that humans do \cite{mitchell2020priorless}, and their difficulty with math and logical reasoning \cite{frieder2023mathematical,borji2023failures} --- things that one might expect a computer to be good at --- suggests that the labyrinthine network of finely tuned parameters is sometimes overlooking simple rules. 

Following the principle of scientific parsimony, we believe a simple explanation is better. Our method, which we call the logical grammar embedding (LGE), is designed to produce an explanation of the data that is clear, compact, and easy to interpret. Grammatical categories are expressed in a discrete algebra such that every word is encoded with a small number of bits of information, and the structure of the bits within the code is connected to the structure of the sentence as a whole. The structure of the algebra makes it easy to interpret what the algorithm has inferred, and the discreteness makes it obvious which words are considered grammatically equivalent. Instead of hundreds of billions of continuous parameters, we have just a handful of bits describing each word in the lexicon.


Invoking scientific principles once more, we posit that a good explanation or theory, in addition to being simple, should be able to make clear predictions about things that have not yet been seen. That is, the explanation that the algorithm learns should allow for the generation of novel output different from the training data. This is a tricky challenge for LLMs: Erring on the side of freedom can lead to the kind of problems that Tay demonstrated; erring on the side of caution leads to an AI unwilling to say anything about interesting or controversial topics \cite{chomsky2023false}. The reporter to whom Sydney professed her love got to see firsthand how an LLM trying to broach a difficult topic can easily stray and trigger its self-censorship protocols \cite{roose2023bing}. The problem is a lack of transparency: If we knew what freedoms were leading to the unwanted behavior, it would be easier to prevent it internally rather than imposing external safety brakes to prevent the algorithm from producing harmful output. In LGE, the process of generating novel output is completely transparent, and the sources of freedom are obvious and controllable.

Finally, one should not need to ingest the entire internet to learn grammar. Indeed, linguists are often quick to point out that humans acquire their linguistic faculties without seeing anywhere near the amount of data that LLMs require \cite{yang2017growth,chomsky2023false}. The strong discrete constraints of LGE enable it to make useful inferences from small data sets --- as few as a hundred sentences, instead of hundreds of gigabytes of training data. A smaller data set is much easier to vet, to make sure it is not tainted with information that would guide the algorithm down a dangerous path, and to identify any biases that may be present.

\section{Method}
\subsection{Logical grammars}
Our method is inspired by two closely related kinds of formal grammar, both of which emphasize the notion that the ordering of the parts of speech in a sentence \new{is not an arbitrary convention; there ought to be some logical basis}. In a categorial grammar (CG) the \new{syntactic category assigned to each word determines which other categories must occur next to it, and the way adjacent categories} are combined parallels the way semantic meaning is processed \cite{steedman1993categorial}. There are two types of categories: A \textit{base type} is meant to be a fundamental entity that is not well expressed in terms of anything else. A classic choice is to take three base types: \textsf{S}, representing a sentence; \textsf{NP}, representing a noun phrase; and \textsf{N}, representing a noun. A \textit{derived type} is a functional type that maps from one (base or derived) type to another. \new{The order of words in a sentence is valid if these functions receive their arguments in the right place}. We use slashes to denote functional arguments, with the direction of the slash indicating whether the function expects its argument on the right or left. Our notation is not necessarily standard, but as an example, we can write an intransitive verb as $\textsf{IV}=/\textsf{NP}\,\textsf{S}$, meaning a function that expects a noun phrase to its left and, if it receives one, returns a sentence. An adjective expects a noun to its right and produces another noun, so it would be \textsf{N} \textsf{N}$\backslash$.

The functional language of a CG is useful for thinking about the role of a word or phrase in a sentence, especially in the context of semantics, but the actual computations of parsing a sentence are often easier to think about in the language of abstract algebra. A pregroup grammar (PG) classifies words not as functional types but as as elements of a pregroup --- a partially-ordered monoid wherein every element $x$ has a left adjoint $x\backslash$ and a right adjoint $/x$ \cite{lambek2008pregroup}. We find that the pregroup and its partial order introduce subtleties that are unnecessary for the computations involved in our algorithm, so we will instead formulate our model in terms of a quasigroup with identity (also known as a loop) \cite{kunen1996quasigroups}. It is a set equipped with
\begin{itemize}
    \item a multiplication operation, which we denote (for example) $x\cdot y$,
    \item an identity $1$ such that $1\cdot x = x = x\cdot 1$ for all $x$, and 
    \item a left inverse $x\backslash$ and right inverse $/x$ such that $x\backslash\cdot x=1=x\cdot /x$.
\end{itemize}
In other words, it is a group without associativity. (Nonassociativity is necessary for $x\backslash$ and $/x$ to be distinct.) Every word is assigned an element of the quasigroup, and a sentence is valid if the words can be multiplied together to produce the identity.

\begin{figure}
    \centering
    \includegraphics[width=\figwidth]{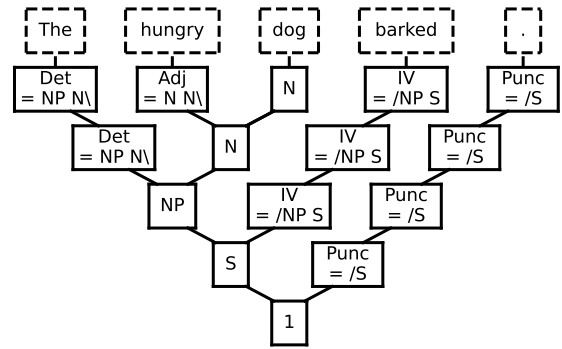}
    \caption[Logical grammar parse tree]{An example of a parse tree. Each node is labeled by the grammatical type present that that node. Here the base types are the sentence \textsf{S}, noun phrase \textsf{NP}, and noun \textsf{N}. The derived types are expressed in terms of the base types, such as the intransitive verb $\textsf{IV}=/\textsf{NP}\,\textsf{S}$. One can think of this as a function that takes an \textsf{NP} on its left as an argument and returns an \textsf{S}. Alternatively, one can think of /\textsf{NP} as the right inverse of \textsf{NP}, so that the product of \textsf{NP} and \textsf{IV} is \textsf{S}.
    }
    \label{fig:tree}
\end{figure}

Figure \ref{fig:tree} is a \textit{parse tree}, an illustration of how to parse a sentence.
We use the classic CG base types \textsf{S} \textsf{NP} \textsf{N}, but we can think of them as \new{generators} of the quasigroup. Likewise, we can think of each branching event as the application of a function, or as the cancellation of an element with its inverse. For example, the determiner $\textsf{Det}=\textsf{NP}\,\textsf{N}\backslash$ contracts with the noun \textsf{N} to produce \textsf{NP}. Thus, the parse tree is a step-by-step derivation of how the product of all the elements in the top layer of the tree can be reduced to the element at the root node. We stipulate that a complete sentence with punctuation should have the identity at its root\new{, which we can accomplish by encoding the punctuation as the right inverse of the sentence type.} This way, the set of properly punctuated grammatical utterances is the set of expressions that are equal to the identity, and the ability to string together any number of valid sentences follows naturally from the fact that the identity times the identity is still the identity. 

As an example of how the algebraic formalism can be simpler than the functional formalism, consider the transitive verb. The CG way of representing a transitive verb would be as a function that receives an \textsf{NP} to its right and returns a function that (farther down the parse tree) receives an \textsf{NP} to its left and returns an \textsf{S}. Symbolically, this would be $(/\textsf{NP}\,\textsf{S})\,\textsf{NP}\backslash$. Apart from being more difficult notation-wise, such higher-order functions also make it more difficult to classify the words in a data set because we need a way to encode all possible derived types, and they become exponentially more numerous for higher-order functions. In the algebraic notation, we can simply write $/\textsf{NP}\,\textsf{S}\,\textsf{NP}\backslash$, reflecting the fact that it needs an \textsf{NP} to its left and to its right in order to produce \textsf{S}, but the order in which the two \textsf{NP}s are supplied in the parse tree does not really matter. This means the algebraic constraints can be weaker than the functional constraints of a CG, but instead of having to express every possible function at a given node in the parse tree, we just need to express the presence or absence of a few base types and their inverses.

\subsection{Syntax vs.\ semantics}
In the established paradigm of LLMs wherein every word is chosen based on the cumulative statistical context of many previous words, there is no separation between syntax and semantics. But humans certainly understand that there is a difference. One is about structure, the other is about meaning. Conflating the two may well be the reason why LLMs ``hallucinate'' things that are completely false but sound reasonable \cite{ji2022hallucination}. It is possible to learn structure from example text alone, but understanding truth and meaning requires interaction with the actual, physical world. Forgetting or obscuring this fact can be dangerous.

In this paper we are not endowing our algorithm with any ability to interact with the world, so we do not want it to pretend to understand meaning. However, the logical structure embedded in our framework makes it possible to implement semantic understanding in a compositional and reliable way. 
If one is able to classify words into functional types like those of a CG, one can evaluate the meaning of a sentence by simply traversing down the parse tree, evaluating the appropriate function at each layer to combine the meanings of adjacent words, until reaching the sentence node \cite{coppack2021semantics}. A PG can provide a similar semantic capability \cite{meichanetzidis2020quantum}. 
Being able to decide if a claim is true or false, and having a clear step-by-step explanation of that decision built into the parse tree, could be an avenue to address the worrying mendacity that today's large language models can display. Conversely, it could also make it easier for a machine to learn from statements that a user makes --- again, with an interpretable step-by-step explanation of what each statement means, so that the source of the machine's factual knowledge is not shrouded in mystery, obscured under a mountain of training data and a web of hundreds of billions of parameters.

\subsection{The logical grammar embedding}
\label{subsec:lge}
Linguists appreciate the structure and interpretive power of logical grammars like the CG and PG, but for computer scientists trying to process natural language it is more natural to encode words in a continuous vector space. We will embed the discrete logical structure that is so useful for explaining concepts in linguistics in a compact vector format that allows for quick computation without sacrificing expressivity and flexibility. Previous work on unsupervised part-of-speech tagging has mostly been based on deriving statistics from a large data set, as in a Markov model \cite{stratos2016markov}, a clustering method \cite{biemann2009cluster}, or a combination of both \cite{he2018invertible}. Efforts to incorporate the structure of a logical grammar into this process have been rare, usually requiring the user to give the algorithm a list of categories to look for \cite{watkinson2000categorial}. The logical grammar embedding (LGE) allows the algorithm to discover the categories it needs and relies on structure rather than statistics, giving it the ability to make inferences from much smaller data sets.

We endow each node of the parse tree with a \textit{category vector} comprising a set of bytes, one for each base type. Each byte contains a number of bits. We can think of one of the bits as representing the base type, and the bit to its right as its left inverse. (It may sound backwards, but we will see shortly why this ordering is useful.) As an example, if we have a three-bit byte for a base type $x$, then we encode the algebraic element $(/x)^a(x)^b(x\backslash)^c$ bitwise as $abc$. In particular, $x$ would be encoded as $010$, $/x\,x$ would be encoded as $110$, and so on. 

This encoding has some obvious limitations. In particular, we can only express elements in which $/x$, $x$, and $x\backslash$ appear in that order. But in practice this is the most natural way to order them to prevent cancellations --- it would be pointless to try to encode $x\,/x\,x\backslash$ as the expression at some node because that would \new{be equal to} $x\backslash$. \new{Our bit} ordering allows us to avoid cancellations within a node \new{and enable} cancellations when two neighboring nodes are multiplied together. For a byte of length $n$ there are $2n$ ways of writing it as a product; e.g., for $n=3$:
\begin{equation}
    \begin{aligned}
    abc &= abc \cdot 000\\
    abc &= ab0 \cdot 00c \quad &abc = ab1 \cdot 01c\\
    abc &= a00 \cdot 0bc \quad &abc = a10 \cdot 1bc\\
    abc &= 000 \cdot abc.\\
    \end{aligned}
    \label{eq:product}
\end{equation}
The two equations on the right are the ones in which a base type and its inverse have been inserted.


Another limitation of this encoding is the number of inverses it can express. If we only have three bits, then we can express $x$ and its left and right inverses. We would need five bits if we also wanted to express the right inverse of the right inverse $//x$ and the left inverse of the left inverse $x\backslash\backslash$. However, we constructed plausible parse trees for some typical sentences and found that such higher-order inverses were unnecessary, which accords with linguists' observations that higher-order adjoints are not often needed in a PG \cite{lambek2008pregroup}.
\begin{table}[t]
    \centering
    \begin{tabular}{l|c|c}
        category & \begin{tabular}{@{}c@{}}algebraic\\expression\end{tabular} & bit encoding\\
        \hline
        sentence & \textsf{S} & $010\;000\;000$\\
        noun phrase & \textsf{NP} & $000\;010\;000$\\
        noun & \textsf{N} & $000\;000\;010$\\
        punctuation & /\textsf{S} & $100\;000\;000$\\
        conjunction & /\textsf{S}\;\textsf{S}\;\textsf{S}$\backslash$ & $111\;000\;000$\\
        intransitive verb & /\textsf{NP}\;\textsf{S} & $010\;100\;000$\\
        transitive verb & /\textsf{NP}\;\textsf{S}\;\textsf{NP}$\backslash$ & $010\;101\;000$\\
        adverb & \textsf{S}\;\textsf{S}$\backslash$ & $011\;000\;000$\\
        determiner & \textsf{NP}\;\textsf{N}$\backslash$ & $000\;010\;001$\\
        adjective & \textsf{N}\;\textsf{N}$\backslash$ & $000\;000\;011$\\
        preposition & /\textsf{N}\;\textsf{N}\;\textsf{NP}$\backslash$ & $000\;001\;110$\\
    \end{tabular}
    \caption[Example category encodings]{A possible bit encoding to explain simple declarative sentences using \textsf{S}, \textsf{NP}, and \textsf{N} as base types.}
    \label{tab:encodingexample}
\end{table}
Table \ref{tab:encodingexample} provides an example of how one might encode the grammatical categories involved in simple declarative sentences. 
Note that, although the order of bits within a byte is important, we do not give any meaning to the order of the bytes. In essence, we are allowing the \new{generators of the quasigroup (the base types)} to commute with each other for the sake of the encoding. This allows us to encode categories like the transitive verb /\textsf{NP} \textsf{S} \textsf{NP}$\backslash$ without having to split apart the \textsf{NP} byte.

\begin{figure}
    \centering
    \includegraphics[width=\figwidth]{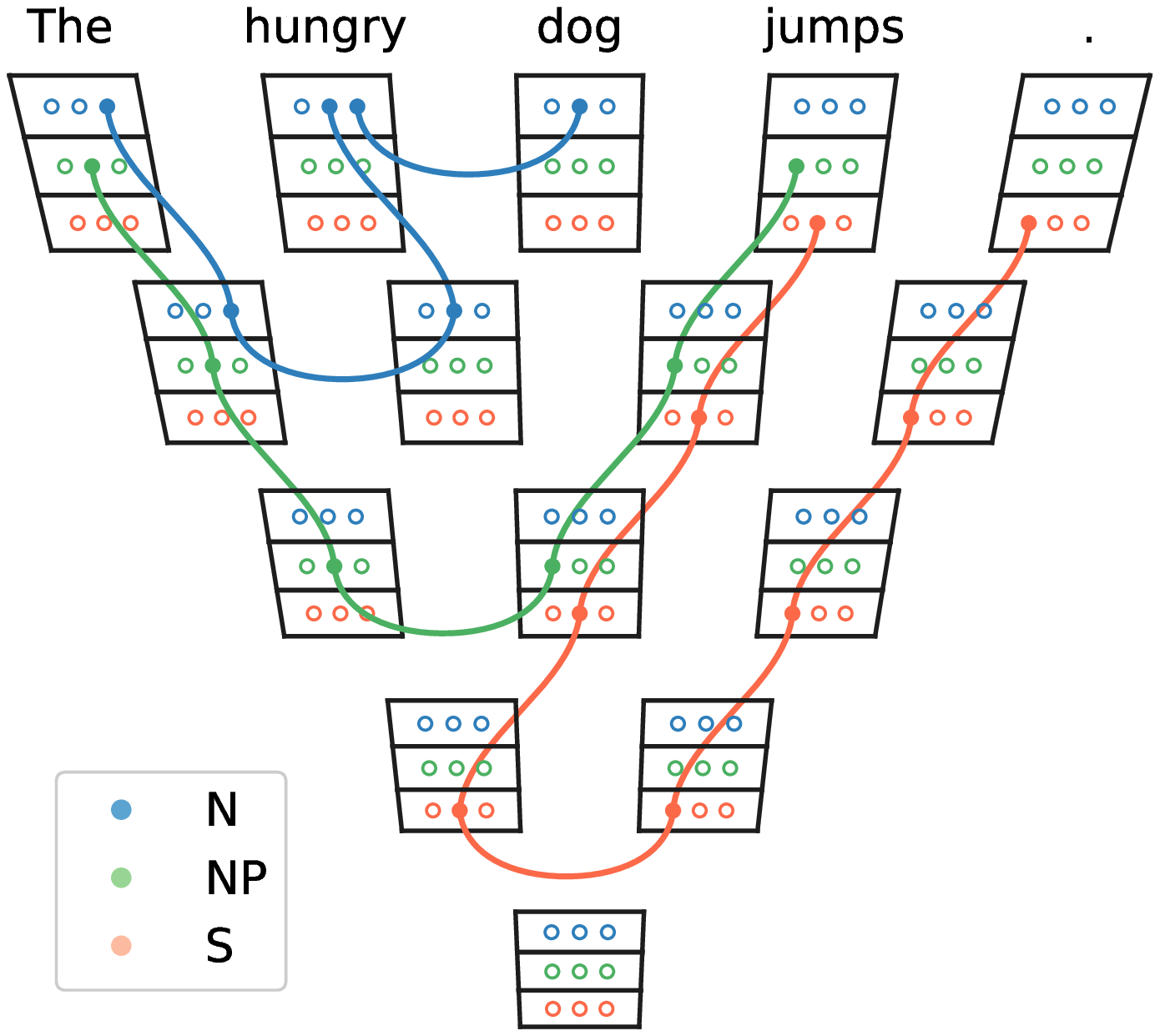}
    \caption[Parse tree illustrating bit cancellations]{
    A parse tree envisioning bit-inverse cancellations as pair-production or annihilation processes. 
    The boxes at each node represent the category vector, with nine bits grouped into three bytes. 
    The filled-in bits are ``on'' --- i.e., the corresponding element of the algebra is present --- and the squiggly lines reflect a ``conservation law'' --- a bit is always preserved from one layer to the next unless it cancels with an inverse. 
    The figure loosely evokes a Feynman diagram with the bits as particles and the squiggly paths as world lines. If the direction of time is toward the bottom of the page, then the process of parsing a sentence is akin to the annihilation of particles with their inverses until none are left. Alternatively, we can reverse the direction of time and use this to visualize the process of generating a novel sentence as a series of pair-production events.
    }
    \label{fig:squiggles}
\end{figure}
Figure \ref{fig:squiggles} illustrates how the cancellations work in a parse tree. The boxes at each node represent the category vector, with nine bits grouped into three bytes. Darker color indicates that a bit is ``on''. The root node of the tree has all bits off, because it must be the identity. 
\new{The fact that the base types commute with each other} is manifested in the image by the fact that the three bytes are separated in the third dimension and do not interact. For readers with color access, the red, green, and blue colors also differentiate the three bytes. 
The lower (red) byte at each node represents \textsf{S}, the middle (green) byte represents \textsf{NP}, and the upper (blue) byte represents \textsf{N}. 
The lines illustrate the fact that a bit is always conserved from one layer to the layer below unless it cancels with one of its inverses. Thanks to our ordering of the bits within each byte these lines do not cross. (If we used the opposite order, we would have a crossing just below the ``hungry'' node.)

The paths traced in Figure \ref{fig:squiggles} suggest squiggly world lines in a Feynman diagram. Indeed, one can think of the parsing process as a series of annihilation events between elements and their inverses. In this way of thinking, the task of our algorithm is to assign a set of ``particles'' to each word such that every sentence can annihilate itself. Conversely, once we have solved that problem, we can also use this picture to envision the process of generating novel sentences as a series of pair production events, starting from the vacuum and culminating in a string of particles that match up with words learned from the training data.
 
The method we present here improves on an earlier effort at interpretable grammar inference \cite{deyo2022grammar} in several ways. The algebra-inspired structure of the category vector builds syntax directly into the categorical encoding, obviating the need for a separate ``syntax tensor'' in the earlier work. Apart from being more elegant and more closely tied to CGs and PGs, this change comes with some technical benefits. It eliminates the need for selecting and tuning several parameters associated with the syntax tensor \cite{deyo2022grammar}, and speeds up the computation time considerably. As a function of grammatical specificity (total number of bits in the category vector), the computational work required at each layer of a parse tree is linear (instead of cubic) in the length of the vector. These improvements allow us to handle a greater size and variety of data without explosive growth in computation time or potentially tedious parameter tuning.

\subsection{Parsing the data}
\label{subsec:parsealgo}
The task is to find a parse tree for every sentence in the data set. Each parse tree has a single node at its base, two nodes in the layer above, then three, and so on until the number of nodes is equal to the number of tokens (including punctuation) in the sentence. The category at the base node must be the identity of the algebra. Every layer must be related to the layer above it by a single branching event: The category vector at one node in the layer expands into a product of two nodes in the layer above, with the category vectors at all other nodes unchanged. Meanwhile, every leaf node of every parse tree is associated with a particular word. All of the nodes associated with a given word should have the same category vector. 
The global nature of this word constraint, acting across all sentences in the data set, is what makes the discovery of parse trees nontrivial.

We employ a ``divide and concur'' strategy \cite{gravel2008divide} to separate the problem into a collection of straightforward computations. These computations constitute projection operations to two sets $A$ and $B$. We then use the ``relaxed-reflect-reflect'' scheme \cite{elser2021learning} to iteratively search for a point in $A\cap B$, which will be our solution. 

Simultaneously finding satisfactory parse trees for all of the data is difficult, but if we just look at a single layer at a time and forget all the others then it is relatively straightforward to ensure that the layer is algebraically equal to the layer above it. Similarly, if we forget about the layer constraints, it is easy to ensure that the leaf nodes associated with a given word agree.

Of course, the difficulty is that each node is involved in more than one of these easy constraints. To overcome this difficulty we give each node as many replicas of its category vector as it needs to satisfy these simple constraints individually, then try to make the replicas concur with each other. Each node (except the root nodes) needs two replicas --- one to be related to layer below, one for the layer above. For the leaf nodes there is no layer above, but the second replica is useful for enforcing the word constraints.

To put things in more mathematical terms, let $v$ be the concatenation of all the replicas of category vectors at all nodes in all parse trees. All of the bits in the vectors ought to be binary, but let us consider $v$ as a point in a high-dimensional continuous Euclidean space. Let $A$ be the set of all $v$ that satisfy
\begin{itemize}
    \item the layer constraints --- each layer is connected to the layer above it by a single (algebraically valid) branching event; and
    \item the word constraints --- all leaf nodes for a given word are equal.
\end{itemize} 
Note that the word constraints do not require the category vectors to be discrete --- the layer constraint will take care of that --- so the word constraint computation consists of averaging all leaves for a given word.

The layer constraint is perhaps the least obvious part of the projection. For every node in a given layer, we compute the distance it would take to project $v$ to a set of binary category vectors where a branching event occurs at that node. This includes making that node equal to the product of the two nodes above it, and making sure all nodes to the left and right are preserved from this layer to the next. Making one node equal to the product of two is a matter of choosing the best option from among Eqns. \eqref{eq:product} for each byte. We then simply choose the node with the smallest distance. The computation is linear in the number of bits per byte, the number of bytes per node, and the number of nodes in the layer. Since the replicas make each layer constraint independent of the others, we can perform this projection independently for every layer of every parse tree (and independently of the word constraints).

Let $B$ be the set of all $v$ for which
\begin{itemize}
    \item the root node is the identity;
    \item the two replicas at each non-root node are equal; and
    \item at each node except the rightmost node in a layer, the sum of the central bits of each byte is equal to $1$.
\end{itemize}
Recall that the central bits of each byte represent the base types or generators of the algebra (the other bits represent the inverses). Having a single base type present in a category vector makes it easier to interpret the category in analogy with a CG, and certainly having at least one base type seems to be the norm when linguists use PGs \cite{lambek2008pregroup}. Requiring the sum of the central bits to be $1$ is the least intrusive way to do this. (Again, this constraint does not need to be discrete because the layer constraints already enforce discreteness.) To perform the projection we simply average the two replicas at each node, and if this results in a sum of central bits different from $1$ then we add an equal amount to each central bit to make the sum equal to $1$. We exclude the root node from this constraint, as we expect it to be the identity (all zeroes). The reason we exclude the rightmost node in every layer from the central bit constraint is that it will ultimately be associated with the sentence-ending punctuation, and we are perfectly happy for that to be encoded as just the right inverse of a base type as in Table \ref{tab:encodingexample}.

We initialize every element of $v$ with a random number between $0$ and $1$ and iterate with the relaxed-reflect-reflect scheme \cite{elser2021learning}: 
\begin{equation}
    v\to \left(1-\frac{\beta}{2}\right)v+\frac{\beta}{2}R_B(R_A(v)),
    \label{eq:iter}
\end{equation}
where $R_A(v)=2P_A(v)-v$ is the reflection of $v$ across $P_A(v)$, the projection of $v$ to $A$; likewise for $B$. The parameter $\beta$ controls how aggressively each iteration updates. Taking \new{$\beta\to1$ yields fast convergence when the constraints are convex. Smaller $\beta$ is slower but more reliable when the constraint sets are nonconvex --- which ours certainly are because of the discreteness.} We set $\beta=0.5$.

One can check that if $v$ is a fixed point of \eqref{eq:iter} then $P_A(v)$ is an element of $A\cap B$, and is therefore a solution to our problem. Natural language is messy, however, so we do not expect exact solutions in general. The change in $v$ in every iteration is roughly proportional to the distance between points in $A$ and $B$ that are near $v$, so we have the algorithm produce its ``best guess'' by taking the concurred category vectors from $P_B(v)$ from whichever iteration $v$ changes the least, and rounding every bit to $0$ or $1$. This way the algorithm can be helpful even when it fails to find an exact solution.

\subsection{Generating new sentences}
\label{subsec:gen}
Once we have parse trees, we can use them to generate new sentences. The generation process is independent of the parsing process, so we have significant freedom regarding how we wish to proceed. The simplest approach is to look through the parse trees that the algorithm has learned and record
\begin{itemize}
    \item the syntactic \textit{branch rules} at each layer of the tree where one code is written as the product of two; and
    \item the lexical \textit{leaf rules} at the top of the tree, where each code is replaced with a single word or token.
\end{itemize}
These rules alone would be \textit{context-free}, in the sense that the rule applied to a node is completely independent of any adjacent nodes. However, we will see in Section \ref{subsec:synthgen} that, though this can be enough to capture the core syntax --- the ordering of nouns, verbs, etc. --- it fails to capture nuances like the difference between subject and object pronouns and the agreement between subject and verb in grammatical number (singular vs.\ plural). We can account for some of these nuances by introducing a small amount of context: When choosing what rule to use for a given code, instead of choosing among all rules that have been used for that code, we look at the code to the left and the code to the right and choose a rule only from among the rules that have been used for that code when it has had the same left and right neighbors. Note that we are only looking at the neighboring codes, not the neighboring words, so this is not an $n$-gram model: The model has the ``creative freedom'' to try unseen word combinations if they are algebraically valid. But again, there is no reason we cannot make a different choice: If we want to restrict the model to imitate the data set more closely, we could allow it to consider neighboring words in addition to codes. Another controllable source of freedom: Some rules will be used more often than others in the parse trees. When deciding what rule to apply at a given node, we can choose from the possible rules with a distribution weighted by the number of uses in the training data, in order to imitate the data more closely, or from a uniform distribution, to favor more novel constructions. In this paper we take the former approach. The algebra determines which constructions are allowable; the probabilities determine which ones are typical.

\new{
A useful metric to characterize the generative model is the \textit{perplexity}, which roughly quantifies how many options the model has to choose from every time it makes a decision. Given the distinct character of the branch and leaf rules, we find it convenient to separate their contributions. The branch perplexity is
\begin{equation}
    \mathcal{P}_b = \exp\left(-\frac{\sum_b \log(p_b)}{N_b}\right),
    \label{eq:ppb}
\end{equation}
where the sum is over branches $b$ with $N_b$ the total number of branches (among the trees that were parsed successfully), and $p_b$ is the probability of the branch rule that was used at $b$ (among the rules that could have been used at that point). The leaf perplexity is then
\begin{equation}
    \mathcal{P}_l = \exp\left(-\frac{\sum_l \log(p_l)}{N_l}\right),
    \label{eq:ppl}
\end{equation}
where the sum is over leaves $l$ with $N_l$ the total number of leaves and $p_l$ the probability of the leaf rule that was used at $l$. We combine them geometrically to obtain a total perplexity
\begin{equation}
    \mathcal{P} = \sqrt{\mathcal{P}_b\,\mathcal{P}_l}.
    \label{eq:pp}
\end{equation}
Now, we must be cautious about reading too much into this metric. 
Lower perplexity is not necessarily better: If the goal is to classify words --- group the nouns together, the verbs together, and so on --- then it might be better to have a low branch perplexity and a high leaf perplexity, because that would mean that there are more words in a small number of categories instead of many categories with only one or two words. Then again, if that imbalance goes too far then we might not be classifying the words with adequate precision.} 
We find that perplexity is most useful for selecting among several solutions (using different random initializations of $v$) to a given data set: If there is a large spread of perplexity values then the solutions with lower perplexity are generally better, though overfitting is still a possibility. 
We believe the most instructive measure of success is to look at the algorithm's generated sentences. Checking that they sound reasonable ensures that the algorithm has classified words with appropriate specificity, and we can guard against overfitting by checking that the sentences are not merely reproductions from the training data.

A repository with code for parsing sentences and generating new ones is available at https://github.com/seandeyo/LGE.

\section{Experiments}
\subsection{Synthetic data as a test case}
\label{subsec:synth}
First we will apply LGE to a synthetic data set of simple declarative sentences, as a sanity check and a demonstration of the algorithm's behavior in a controlled setting. The rules we use to generate the data are given in Table \ref{tab:synthrules}. We start with the identity \textsf{1} and choose randomly from among the available rules until we have a string of words with a period at the end. 
Table \ref{tab:syntheg} gives ten example sentences. 

\begin{table}[t]
    \centering
    \begin{tabular}{rcl}
        category & & options \\
        \hline
        \textsf{1} & $\to$ & \textsf{S}\;\textit{.} $|$ \textsf{S}\;\textsf{Conj}\;\textsf{S}\;\textit{.} \\
        \textsf{S} & $\to$ & $\textsf{Subj}_\text{s,p}\;\textsf{VP}_\text{s,p}$ $|$ $\textsf{Subj}_\text{s,p}\;\textsf{Adv}\;\textsf{VP}_\text{s,p}$ \\
        $\textsf{Subj}_\text{s}$ & $\to$ & \textit{she} $|$ \textit{he} $|$ \textit{it} $|$ $\textsf{Det}_\text{s}\;\textsf{N}_\text{s}$ $|$ $\textsf{Det}_\text{s}\;\textsf{Adj}\;\textsf{N}_\text{s}$ \\
        $\textsf{Subj}_\text{p}$ & $\to$ & \textit{they} $|$ $\textsf{Det}_\text{p}\;\textsf{N}_\text{p}$ $|$ $\textsf{Det}_\text{p}\;\textsf{Adj}\;\textsf{N}_\text{p}$ \\
        $\textsf{VP}_\text{s,p}$ & $\to$ & $\textsf{IV}_\text{s,p}$ $|$ $\textsf{VP}_\text{s,p}\;\textsf{Obj}$ \\
        \textsf{Obj} & $\to$ & \textit{her} $|$ \textit{him} $|$ \textit{it} $|$ \textit{they} $|$ \\
        & & $\textsf{Det}_\text{s,p}\;\textsf{N}_\text{s,p}$ $|$ $\textsf{Det}_\text{s,p}\;\textsf{Adj}\;\textsf{N}_\text{s,p}$ \\
        $\textsf{N}_\text{s}$ & $\to$ & \textit{human} $|$ \textit{dog} $|$ \textit{bear} $|$ $\textsf{N}_\text{s}\;\textsf{Prep}\;\textsf{NP}$ \\
        $\textsf{N}_\text{p}$ & $\to$ & \textit{humans} $|$ \textit{dogs} $|$ \textit{bears} $|$ $\textsf{N}_\text{p}\;\textsf{Prep}\;\textsf{NP}$ \\
        $\textsf{IV}_\text{s}$ & $\to$ & \textit{runs} $|$ \textit{jumps} $|$ \textit{sits} $|$ \textit{speaks} \\
        $\textsf{IV}_\text{p}$ & $\to$ & \textit{run} $|$ \textit{jump} $|$ \textit{sit} $|$ \textit{speak} \\
        $\textsf{TV}_\text{s}$ & $\to$ & \textit{sees} $|$ \textit{hears} $|$ \textit{follows} $|$ \textit{avoids} \\
        $\textsf{TV}_\text{p}$ & $\to$ & \textit{see} $|$ \textit{hear} $|$ \textit{follow} $|$ \textit{avoid} \\
        \textsf{Conj} & $\to$ & \textit{and} $|$ \textit{but} $|$ \textit{until} $|$ \textit{while} \\
        \textsf{Adv} & $\to$ & \textit{always} $|$ \textit{often} $|$ \textit{seldom} $|$ \textit{never} \\
        \textsf{Prep} & $\to$ & \textit{by} $|$ \textit{with} $|$ \textit{near} $|$ \textit{beside} \\
    \end{tabular}
    \caption[Rules used to generate synthetic data]{Rules used to generate sentences in the synthetic data set. When a category has more than one possible rule, a vertical bar $|$ separates the various options. The subscripts \textsf{s} and \textsf{p} are shorthand for singular and plural. The combination \textsf{s,p} means either is allowed but it must match: For instance $\textsf{S}\,\to\,\textsf{Subj}_\text{s}\,\textsf{VP}_\text{s}$ is legal but $\textsf{S}\,\to\,\textsf{Subj}_\text{s}\,\textsf{VP}_\text{p}$ is not.}
    \label{tab:synthrules}
\end{table}

\begin{table}[t]
    \centering
    \begin{tabular}{l}
        \hline
        \textit{they jump until a human sits .}\\
        \textit{all bears near a bear sit .}\\
        \textit{he speaks .}\\
        \textit{all dogs near some happy bears avoid a bear .}\\
        \textit{he jumps while all bears seldom speak .}\\
        \textit{they hear them .}\\
        \textit{all humans seldom run .}\\
        \textit{they always see some sad dogs while it speaks .}\\
        \textit{some happy bears jump but the bear sits .}\\
        \textit{a bear always speaks until he follows all humans .}\\
        \hline
    \end{tabular}
    \caption[Examples of synthetically generated sentences]{Example sentences generated by the rules in Table \ref{tab:synthrules}.}
    \label{tab:syntheg}
\end{table}

\begin{figure}
    \centering
    \includegraphics[width=\figwidth]{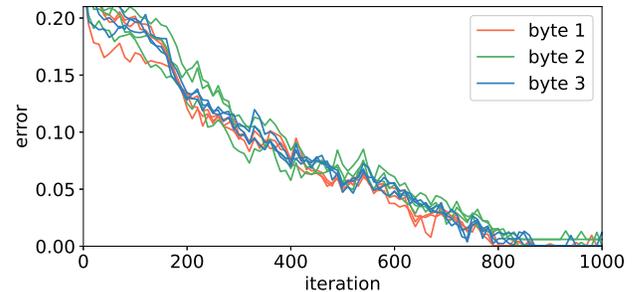}
    \caption[Error vs.\ iteration number, training on synthetic data]{Error vs.\ iteration number for one trial as the algorithm tries to parse 100 sentences from our synthetic data set. The nine different curves show the error separately for each bit, with the three bits for each byte plotted in the same color.}
    \label{fig:syntherror}
\end{figure}
Figure \ref{fig:syntherror} shows how the algorithm's error evolves as it tries to parse 100 of these sentences. By ``error'' we mean how much the variable $v$ changes in each iteration of \eqref{eq:iter}. Each of the curves in the figure is the rms error for a single bit over all category vectors in all parse trees. We plot bits of the same byte in the same color. The fact that the bits contribute more or less equally to the error suggests that the work of parsing is being distributed more or less equally among the bits. 

\begin{table}[t]
    \centering
    \begin{tabular}{l|l}
        code & words \\
        \hline
        010 000 000 & \textit{they} $|$ \textit{he} $|$ \textit{them} $|$ \textit{it} $|$ \textit{she} $|$ \textit{her} $|$ \textit{him}\\
        100 010 000 & \textit{jump} $|$ \textit{sits} $|$ \textit{sit} $|$ \textit{speaks} $|$ \textit{jumps} $|$ \textit{speak} $|$\\
        & \textit{run} $|$ \textit{runs}\\
        000 111 000 & \textit{until} $|$ \textit{while} $|$ \textit{but} $|$ \textit{and}\\
        000 000 010 & \textit{a} $|$ \textit{all} $|$ \textit{some} $|$ \textit{the}\\
        010 000 100 & \textit{human} $|$ \textit{bears} $|$ \textit{bear} $|$ \textit{dogs} $|$ \textit{humans} $|$ \textit{dog}\\
        000 100 000 & \textit{.}\\
        100 011 000 & \textit{near} $|$ \textit{beside}\\
        011 000 000 & \textit{happy} $|$ \textit{hungry}\\
        101 010 000 & \textit{avoid} $|$ \textit{hear} $|$ \textit{see} $|$ \textit{follows} $|$ \textit{follow} $|$ \textit{hears} $|$\\
        & \textit{avoids} $|$ \textit{sees}\\
        110 000 000 & \textit{seldom} $|$ \textit{often}\\
        000 011 000 & \textit{always} $|$ \textit{never}\\
        000 000 110 & \textit{sad}\\
        100 000 011 & \textit{by}\\
        111 000 000 & \textit{with}\\
    \end{tabular}
    \caption[Example of word categories inferred from synthetic sentences]{An example of the word categories learned by the algorithm from 100 sentences generated according to Table \ref{tab:synthrules}.}
    \label{tab:synthsol}
\end{table}
Table \ref{tab:synthsol} gives an example word encoding that the algorithm learned from 100 sentences generated according to Table \ref{tab:synthrules}. They mostly line up with our expectations --- especially the first two bytes, which seem to be playing the role of \textsf{NP} and \textsf{S}, respectively. The codes can also reveal some symmetries of the grammar. Linguists prefer to use \textsf{N} as the the third base type, but in this example the algorithm seems to be using its third base type for determiners, an equally valid choice.
Another symmetry is that a word can often be encoded just as easily as the left inverse of something to its right or the right inverse of something to its left. In our synthetic sentences, if an adverb is present it is always after a noun phrase and before a verb. Consequently, it is possible to encode the adverbs as noun phrase right-modifiers /\textsf{NP} \textsf{NP} (as with \textit{seldom} and \textit{often} in this example) or as sentence left-modifiers \textsf{S} \textsf{S}$\backslash$ (\textit{always} and \textit{never}). \new{Of course, this is only one trial: Different random initializations of $v$ yield different solutions. For this data set the differences are usually symmetries like the ones just described and they mostly affect adverbs, adjectives, and prepositions.}


\subsection{Generating sentences}
\label{subsec:synthgen}

\begin{table}[t]
    \centering
    \begin{tabular}{r}
        \hline
        \textit{a human hear some sad dogs .}\\
        \textit{all sad bear see some sad happy bear .}\\
        \textit{all sad bear run .}\\
        \textit{they always runs .}\\
        \textit{he often jump .}\\
        \textit{a humans jump .}\\
        \textit{some bear follows they .}\\
        \textit{some humans speaks .}\\
        \textit{a human jumps until all sad hungry bear follows they .}\\
        \textit{the dogs speaks .}\\
        \hline
        \textit{the happy dogs speak .}\\
        \textit{some sad dogs avoid it .}\\
        \textit{the hungry human speaks .}\\
        \textit{they follows all humans .}\\
        \textit{all sad dogs seldom run until the human seldom jump .}\\
        \textit{some dogs near the hungry dogs avoid all hungry bear .}\\
        \textit{the bear near the hungry dogs follow the humans .}\\
        \textit{some dogs with some happy bears seldom speak .}\\
        \textit{some bears seldom run until they follows it .}\\
        \textit{she follows them .}\\
        \hline
    \end{tabular}
    \caption[Sentences generated after training on synthetic data]{Sentences generated after training on 100 sentences from the synthetic data. We take the branching events of the learned parse trees as production rules to generate new sentences. Top: If we use the rules in a context-free fashion (only considering the bits at one node when deciding what rule to apply at that node), the basic structure is right but there is no distinction between subjects and objects (e.g., \textit{they} vs.\ \textit{them}) and the subject and verb often disagree in grammatical number (singular vs.\ plural). Bottom: A minimal amount of context --- considering a node's left and right neighbors when choosing a rule --- corrects the subject-object problem.}
    \label{tab:synthgen}
\end{table}
With parse trees in hand, we can turn things around to generate new sentences as described in Section \ref{subsec:gen}. Table \ref{tab:synthgen} provides some examples. (We align the generated sentences to the right, to help the reader distinguish them from the left-aligned training sentences like those in Table \ref{tab:syntheg}.) If we use the context-free approach, viewing each branch rule or leaf rule as independent of the neighboring codes, the generated sentences get the basic structure right but fail in a few predictable ways (top half of Table \ref{tab:synthgen}). One is subject-object distinction (\textit{they} vs.\ \textit{them}). There are many noun phrases or pronouns that can appear as the subject or object (e.g., \textit{the bear}, \textit{all dogs}, \textit{it}), so it is natural for subjects and objects to have the same code. Another failure, disagreement of grammatical number (singular vs.\ plural), can also be understood by considering transitive verbs: Two sentences with the same subject and verb can differ in their object --- one can have a singular object; the other, plural. Both must be parsed as equal to the identity, so the only choice is to make singular noun phrases equal to plural ones. (It is still possible distinguish between singular and plural for nouns and determiners, however, and some solutions do manage to separate them.)

The usual way to handle these finer features of grammar is to make the categories more complicated. Indeed, we did just that in formulating the production rules of our synthetic grammar in Table \ref{tab:synthrules} by having separate categories for subject and object. We can encourage the algorithm imitate this by allowing larger category vectors (more bytes). Since some words can play more than one role --- e.g., \textit{it} can be a subject or an object --- we can also relax the leaf node average-by-word constraint to a clustering constraint. Each word would be given some number of clusters, its leaf nodes partitioned (by the k-means method), and the leaves within each partition averaged. \new{That way some instances of the word could use one category vectors and others could fall into a different category. We will explore this option in Section \ref{subsec:relax}, but let us first consider something easier.}

In the spirit of simplicity and efficiency, before trying to find parse trees with more complicated codes we can extract more information from the parse trees we have. We make the production process slightly context-sensitive by considering the neighboring codes for each branch or leaf rule (lower half of Table \ref{tab:synthgen}), as described in Section \ref{subsec:gen}. That generally resolves the subject-object problem. Context can also help with singular-plural agreement, though with the limited scope of our context sensitivity this only works when the verb occurs immediately next to a noun (not a pronoun). The sentences will only become more nuanced when we introduce non-synthetic data, so we will stick with context from here on, but we stress that other more sophisticated models including much more context can and should be part of the generation process for practical applications like chatbots. Having context across multiple sentences is certainly useful when crafting a paragraph, for example. But having a system like ours to provide the underlying syntax structure would make it much easier for such a chatbot to know what it is saying. Indeed, the structure that our context-free generated sentences picked up on is precisely the structure that is important for semantic evaluation --- whether a word is singular or plural is not as important as knowing what kind of function the word is, because evaluating functions is how one determines the truth value of a statement in a CG.

\subsection{Natural data}
\label{subsec:parsecoca}
We will use the Corpus of Contemporary American English (COCA) as a source of natural data. The more than one billion words in COCA come from a variety of sources including academic papers, works of fiction, TV and movie subtitles, blogs, and more, over a time span of 1990-2019 \cite{coca}. We will focus on the fiction subset, simply because it seems to contain a large proportion of what we consider ordinary grammatical sentences (fewer sentence fragments and typographical errors than in the TV and movie subtitles, less jargon than in academic papers). The algorithm's output is only as good as its training data, and since it does not need much data we may as well be picky. The COCA website provides free samples to download. These samples come with some redactions, so we remove any sentences that are partially redacted. We know that the presence of a word in multiple sentences is what makes the task of finding parse trees simultaneously for all sentences nontrivial, so we sort the sentences to favor those with common words. Specifically, we count the number of uses for each word in the data set, score each sentence with the median of the counts for each of its words, and sort the sentences by this score. 
We also discard sentences with three or fewer words or with commas, as most of these turn out to be nonsensical sentence fragments. LGE is perfectly capable of parsing short sentences (an interjection can be given the same type as \textsf{S}, for instance) but parsing something with so little structure is not impressive. The effect of leaving these things in the training data is to pollute the generated sentences with fragments, which then makes it more difficult to evaluate whether the algorithm has done a good job recognizing proper grammar. In any case,
Table \ref{tab:cocaeg} shows ten sentences from the top $200$ of our COCA training sentences. Note that COCA separates punctuation from words by a space, and also separate contractions like \textit{wasn't} $\to$ \textit{was n't}. 
\begin{table}[t]
    \centering
    \begin{tabular}{l}
        \hline
        \textit{I want to be a frog !}\\
        \textit{Right in the middle of the night .}\\
        \textit{I know the feeling .}\\
        \textit{That was the way of the world .}\\
        \textit{It was and he was n't ?}\\
        \textit{I say to Savo .}\\
        \textit{I 'm to blame .}\\
        \textit{I was n't a friend of Mitch 's .}\\
        \textit{I wish to God I had .}\\
        \textit{Be a man of the will .}\\
        \hline
    \end{tabular}
    \caption[Sentences from a natural data set]{Ten sentences from the COCA fiction data set.}
    \label{tab:cocaeg}
\end{table}

\begin{figure}
    \centering
    \includegraphics[width=\figwidth]{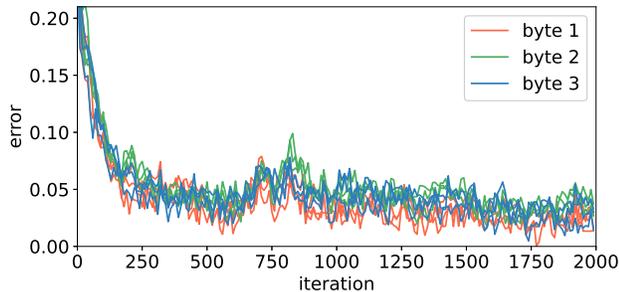}
    \caption[Error vs.\ iteration number, training on natural data]{Error vs.\ iteration number for one trial as the algorithm tries to parse 100 sentences from the COCA fiction data set. }
    \label{fig:cocaerror}
\end{figure}

Figure \ref{fig:cocaerror} shows the error vs.\ iteration number as the algorithm tries to parse $100$ of these sentences. The behavior is similar to that of Figure \ref{fig:syntherror} except that the ``floor'' error that the search eventually reaches is higher, reflecting the messiness of the natural data.
\begin{table}[t]
    \centering
    \begin{tabular}{r}
        \hline
        \textit{I was visiting a friend of the workers !}\\
        \textit{I went to the beginning .}\\
        \textit{I was n't .}\\
        \textit{That was a friend of the Blefuscians .}\\
        \textit{I was referring to the truck .}\\
        \textit{I was a frog !}\\
        \textit{I was a lousy had .}\\
        \textit{* I have to go to orchestra .}\\
        \textit{Easy and a half .}\\
        \textit{I want to be the Head laugh .}\\
        \hline
        \textit{* I get to record the results .}\\
        \textit{Back to the wedding .}\\
        \textit{I was in the clothes of a hurry .}\\
        \textit{I the middle of a lousy bouncer .}\\
        \textit{All was the above .}\\
        \textit{Get to the kitchen .}\\
        \textit{Welcome was a captive and a kamebara .}\\
        \textit{* Back to the sentence .}\\
        \textit{I pointed to the thunderbolt .}\\
        \textit{I showed the ax .}\\
        \hline
    \end{tabular}
    \caption[Sentences generated after training on natural data]{Example sentences generated by the algorithm after training on sentences from the COCA data set. Top: Trained on 100 sentences. Bottom: Trained on 200 sentences.
    Recognizing patterns without overfitting is more of a challenge with natural data: The sentences marked with an asterisk indicate that the algorithm has reproduce a line from its training data. 
    }
    \label{tab:cocagen}
\end{table}
Table \ref{tab:cocagen} shows ten sentences generated by the algorithm after training on $100$ (top half of the table) or $200$ (bottom) COCA sentences. Many of them start with \textit{I}, which is reasonable given that many of the training sentences do. \new{Doubling the number of training sentences does not affect the syntax, but it does expand the lexicon: About $10-15\%$ of the generated sentences are reproductions from the training data (with the reproduction rate varying from one solution to the next) after training on 100 sentences, and this is partly due to the limited vocabulary --- reproductions happen about half as often after training on 200 sentences.} We hope the reader will agree that most of these sentences are grammatically plausible, if semantically odd. But that is precisely the point: A sentence can be syntactically sensible even if its meaning is bizarre. Our goal in this work is to learn syntax, not semantics, and our simple generative process was chosen to make this behavior possible. We want to see evidence that the algorithm could do what humans can --- learn syntax from a small data set, and demonstrate that understanding by saying things that are syntactically valid regardless of semantic sensibility.




\subsection{Lexicon inferred from natural data}
The vocabulary of the natural data is, of course, much larger than in our synthetic data.
Instead of printing a large table with the entire lexicon we will highlight a few features.

\new{The inferred base types do not exactly line up with the classic choice of \textsf{S} \textsf{NP} \textsf{N}. Indeed, we have already seen with the synthetic data that it can be just as effective to use the determiner as a base type instead of the noun. The most notable trend in the inferred categories from the natural data is that the most common words are most likely to be encoded as base types (meaning they have binary codes of the form $010\,000\,000$, or $000\,010\,000$, or $000\,000\,010$). This happens quite consistently with the pronoun \textit{I} (recall that such pronouns are essentially equivalent to \textsf{NP}), the determiners \textit{the} and \textit{a}, and the preposition \textit{to}. But if these three categories use up the three base types, what base type is left to play the role of \textsf{S}? We can look at the parse trees to find the answer: If the noun phrase/pronoun \textit{I} is encoded as $010\,000\,000$ then the branching event at the root of a parse tree is often of the form 
\[000\,000\,000\to010\,000\,000\cdot100\,000\,000.\] This means the algorithm is ``overloading'' one of the types to act as both noun phrase and sentence.

This conflation might initially seem concerning, but it actually makes some sense. For one thing, it makes it easier to parse sentence fragments that consist of a noun phrase followed by a period or exclamation mark. (This means the period and exclamation mark are encoded as the right inverse of the noun phrase/sentence type. The question mark is often encoded as the right inverse of some other base type, which makes sense given that questions tend to have a different structure than statements and imperatives.) The concern is that our generated sentences could contain many such fragments, but in practice the use of context in the sentence-generation process allows the model to distinguish between sentences and noun phrases. It can still produce some fragments, but at a rate consistent with the rate of fragments in the training data. In fact, in some solutions it even places the determiners \textit{the} and \textit{a} in the same category as \textit{I} but still manages to produce proper declarative sentences thanks to context.

Of course, we can allocate a fourth or fifth byte to let the algorithm identify more base types. This tends to reduce the overloading phenomenon, which is useful if one wants to interpret the codes directly without parse tree context.

Beyond the base types, the lexicalization is less tidy than with the synthetic data. Thanks to symmetries like those we have already mentioned, plus the fact that many words or syntactic constructions only appear once in the training data, plenty of one-offs make their way into the solution. The total number of categories used is typically around $80$, but most only contain one or two rare words and differ from each other by symmetries. The categories with the most distinct words typically contain what we as English speakers recognize as nouns. Here is the largest category from one solution: 
\begin{quote}
\begin{tabular}{l|l}
    110 100 000 & \textit{cake} $|$ \textit{saloon} $|$ \textit{movies} $|$ \\
    & \textit{Blefuscians} $|$ \textit{half} $|$ \textit{dogs} $|$ \textit{boat} $|$ \\
    & \textit{morning} $|$ \textit{backseat} $|$ \textit{nutshell} $|$ \\
    & \textit{bombings} $|$ \textit{neighborhood} $|$ \textit{frog} $|$ \\
    & \textit{Mitch} $|$ \textit{snow} $|$ \textit{workers} $|$ \textit{children} $|$ \\
    & \textit{atom} $|$ \textit{musician} $|$ \textit{card} $|$ \textit{rebel} $|$ \\
    & \textit{businessman} $|$ \textit{freak} $|$ \\
    & \textit{postnasciturus} $|$ \textit{message} $|$ \textit{news} $|$ \\
    & \textit{music} $|$ \textit{pistol} $|$ \textit{children} $|$ \textit{year}\\
\end{tabular}
\end{quote}
There were a few other large noun-containing categories with codes $110\,000\,001$ and $110\,000\,100$. (This solution was one in which the first byte was used both for the pronoun \textit{I} and the determiners \textit{the} and \textit{a}, leaving considerable room for flexibility in the other two bytes. The preposition \textit{to} used the code $000\,000\,010$ while \textit{in} and \textit{of} used $000\,010\,000$.) Verbs tend to collect in more medium-sized categories, such as these transitive verbs:
\begin{quote}
\begin{tabular}{l|l}
    111 000 000 & \textit{look} $|$ \textit{think} $|$ \textit{involved} $|$ \textit{thought} $|$ \\
    & \textit{killed} $|$ \textit{took} $|$ \textit{became} $|$ \textit{pressed} \\
\end{tabular}
\end{quote}
Some prepositional verbs (ones that appear before prepositional phrases in the training sentences) occurred with the types $110\,001\,000$ or $110\,000\,001$.
One last pattern that stands out to us is that capitalized words at the start of fragments or imperatives also tend to gather together:
\begin{quote}
\begin{tabular}{l|l}
    011 001 001 & \textit{Listen} $|$ \textit{All} $|$ \textit{Be} \\
    001 000 011 & \textit{Tending} $|$ \textit{Want} $|$ \textit{Back}
\end{tabular}
\end{quote}
}

\subsection{Not parsing everything}
\label{subsec:parsefail}
Real-world data is more difficult to parse than our synthetic data. Indeed, the algorithm generally does not reach an exact solution (at least not within a few thousand iterations). But it does output its ``best guess'' parse trees at the end of its allotted iterations, so we can check which sentences fail to parse most often and use this as a marker of items in the data set worthy of a closer look. The sentences that fail most often can reveal typos, as in this example of misplaced punctuation:
\training{
    \textit{. like the wind in a meadow .}
}
This can help us with our selection of training data, as we would prefer to avoid training on such nonsense, 
but the fact that algorithm can identify much of the nonsense that slips through is an encouraging sign of robustness.
Some complaints are a bit more pedantic: The algorithm often 
seems to echo an English teacher's admonition about ending a sentence with a preposition, failing to parse things like
\training{
    \textit{I was beginning to think th was nobody in .\\
    What was it you were speaking of ?}
}
(The typo \textit{th} is from the data, not us.)

Other times the algorithm draws our attention to something different. In our 100-sentence training sample, \textit{friend} is one of the few nouns that appears in multiple sentences, and the algorithm often fails to parse these sentences:
\training{
    \textit{He was a friend of a friend .\\
    I was visiting a friend .\\
    I was n't a friend of Mitch 's .\\
    Once he was a friend of the workers .}
}
In this case the sentences seem perfectly reasonable, desirable even --- having words appear in multiple sentences is what makes the task of parsing nontrivial, and that is what we want. Of course, nontrivial means more difficult, so it is natural that the algorithm would tend to avoid the challenge and focus on the other sentences. One way to force the algorithm to address the difficulty is to add more sentences with \textit{friend}. In fact, if we just add a second copy of the same four sentences to the 100-sentence sample, plus the next four \textit{friend} sentences from the training data for good measure, and run the algorithm again, it usually manages to parse them. Some examples from the generated sentences:
\generated{
    \textit{I see that someone was a friend of a frog !\\
    I thought I was a friend .\\
    I want to be a friend of a nutshell .\\
    I 'm in a friend .\\
    It was a friend .\\}
    
    
}
Evidently having lots of \textit{friend}s is a good thing. More generally, seeing a word multiple times is helpful for learning how to use it, which is exactly why we sorted the training sentences by word popularity, but we will see shortly that for many words a single exposure is enough.

\subsection{Understanding surprises}
Let us examine some more generated sentences as an opportunity to highlight how easy it is to understand surprising behavior when the model is transparent and the data set is small. One of the most salient surprises is when the algorithm uses an unfamiliar word, as in
\generated{
    \textit{I was referring to the Loup-garou .}
}
or
\generated{
    \textit{That was a friend of the Blefuscians .}
}
Naturally, we wonder what these words mean and whether the algorithm is using them correctly. Semantic meaning is beyond the scope of this work, but we can at least interpret the syntax by checking the training data. The words \textit{Blefuscians} and \textit{Loup-garou} appear exactly once each:
\training{
    \textit{I am the king of the Blefuscians !\\
    Back to the Loup-garou .}
}
In both cases, the word in question is used as a noun in the training data, and the usage in the generated sentences appears to be consistent. (For the interested reader, \textit{Blefuscians} is from Johnathan Swift's \textit{Gulliver's Travels} and \textit{Loup-garou} is a term for a werewolf.) It is worth noting that both words appear only once in the algorithm's training sentences, so it has learned how to use them from a single exposure. 

Indeed, this is a general observation: It is perfectly fine if most words appear only once, as long as a few key words show up multiple times, such as \textit{I} and \textit{the}. The proximity of these common words to the less common ones allows the algorithm to infer the role of the latter. And this is not solely a property of nouns. One generated sentence that surprised the authors was
\generated{
    \textit{What about the Mackeys and the \\
    Jacksonville children .}
}
The surprise was that \textit{Jacksonville} was used a modifier. Again, we can look at the training data to understand why. \textit{Jacksonville} appears exactly once, and it appears as a modifier: 
\training{
    \textit{I pointed to the Jacksonville paper .}
}
Another mystery solved. A similar example:
\generated{
    \textit{I was n't fit to be a sudden .}
}
Why is the algorithm using the adjective ``sudden'' as a noun? Because the word only appears once in the sample of training sentences:
\training{
    \textit{It was all of a sudden .}
}
The idiomatic ``all of a sudden'' does indeed make ``sudden'' seem like a noun. If this is the algorithm's only exposure to that word, can we really blame it for this mistake?

Resolving these ``surprises'' may seem completely trivial, but that is the point. When one can make inferences from a small data set, it is easier to identify the sources of surprises. Contrast this with the difficulty of deciphering the choices of an LLM --- e.g., why did Microsoft's Sydney use the purple devil emoji after stating ``I want to be alive''? The answer, all too often, is a shrugging emoji. 

\subsection{Relaxing constraints}
\label{subsec:relax}

Part of what makes natural data more difficult to parse is the diversity of syntactic constructions. When we ask the algorithm to parse, say, $1000$ sentence from the COCA set, it often fails to find valid parse trees for several hundred. \new{Some of these sentences may be grammatically incorrect, but in many cases the tight constraints that worked on the synthetic data are too strict to explain the syntactic diversity of the natural data. Let us consider some ways of relaxing these constraints.

\begin{figure}
    \centering
    \includegraphics[width=\figwidth]{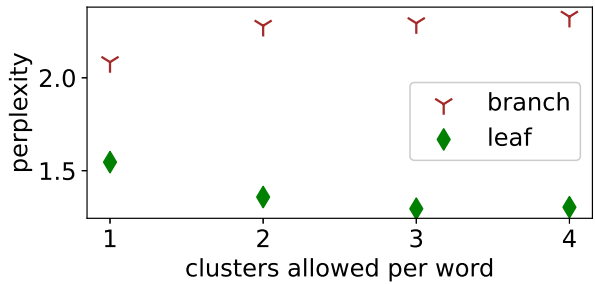}
    \caption[Perplexity vs.\ number of word clusters]{Perplexity vs.\ number of clusters allowed for each word. Here we separate the perplexity into its branch and leaf components. These results are for training on 100 sentences from the COCA fiction data set.}
    \label{fig:clusters}
\end{figure}
First, we recognize existence of homographs --- words that have the same spelling but different meaning. Many of these words can function as nouns or verbs (\textit{start}, \textit{run}, \textit{finish}, etc.). Others may have the same classical part of speech but still play different functional roles --- prepositions would likely need different functional types depending on whether they follow nouns (e.g., \textit{dog in the house}) or verbs (\textit{jump in the water}). To account for this phenomenon, we can relax the word concur constraint: Instead of forcing all leaf nodes associated with a given word to be equal, we can partition them into some number of clusters and require the leaves within a cluster to be equal. Figure \ref{fig:clusters} shows how the balance of perplexity shifts when clusters are allowed: More clusters generally leads to more lexical types with fewer words per type, which means that the branch perplexity increases and the leaf perplexity decreases. (It is often helpful to add a fourth or perhaps fifth byte to provide room for these additional types, especially if one wants to avoid the overloading phenomenon discussed in Section \ref{subsec:parsecoca}.) The greatest change occurs from one cluster (i.e., the strict word concur) to two, with additional clusters making little difference. This suggests that many words can make use of a second cluster (the noun/verb and preposition examples give us an idea of why) but few require more than that.
}


\new{Another relaxation we can make is to allow more than one base type to be present in each category vector. Our enforcement of a single base type at each node (except the root and punctuation nodes) has the effect of requiring exactly one pair production event at each branching point, so we can also think of this relaxation as allowing multiple (or zero) pair productions at each branching point. The relaxation also increases the number of types a given number of bytes can express: Three bytes without this restriction can achieve comparable perplexity to four or five bytes with the restriction. The cost we pay for this efficiency is that it is marginally more tedious to interpret a code with multiple base types, but words with multiple base types do appear in some constructions in PGs, so this may simply be the cost of dealing with syntactic diversity.

One more option is to ``fudge'' the algebra by allowing one of the category vectors produced by each branching event to flip a bit. In the particle language of Figure \ref{fig:squiggles}, this means we are allowing particles to be created or destroyed at branching events. This relaxation is very effective at increasing the parse success rate: When training on $1000$ sentences, the bit flip relaxation can reduce the parse failures from several hundred sentences to a few tens. This relaxation works best in combination with the multiple-base-type relaxation --- there are not as many useful bit flips when one can only have one base type. Table \ref{tab:cocanoise} shows how this combination can extract a variety of syntactic constructions from its training sentences while managing to avoid verbatim reproductions.
}

\begin{table}[t]
    \centering
    \begin{tabular}{r}
        \hline
        \textit{I have a study .}\\
        \textit{He was a dream of a friend .}\\
        \textit{That was her chimney .}\\
        \textit{Get in her service ?}\\
        \textit{This was a sacrifice .}\\
        \textit{There 's do it .}\\
        \textit{What was she song .}\\
        \textit{I requested something to puke that ?}\\
        \textit{How was the wheelbarrow .}\\
        \textit{There was a spare .}\\
        \hline
    \end{tabular}
    \caption[Sentences generated after training on natural data, with allowances for noise]{Example sentences generated by a solution learned from $1000$ sentences of the COCA fiction data set with two relaxations to improve performance on diverse or noisy data: 1) allowing more than one base type in the category vectors, and 2) allowing bit flips in the branching events.}
    \label{tab:cocanoise}
\end{table}

\subsection{Seeding}
\label{subsec:seed}
One convenient method for expediting the parsing process is to use word codes learned from a previous solution as a seed for a new solution. The modification is simple: During the algorithm's concur procedure (the projection to set $B$), for any word whose code we already know, we require the leaf nodes for that word to be equal to the known code. For all other words the concur proceeds as usual. The extra constraints reduce the search space and encourage the algorithm to incorporate new vocabulary into the existing grammar. 

\begin{figure}
    \centering
    \includegraphics[width=\figwidth]{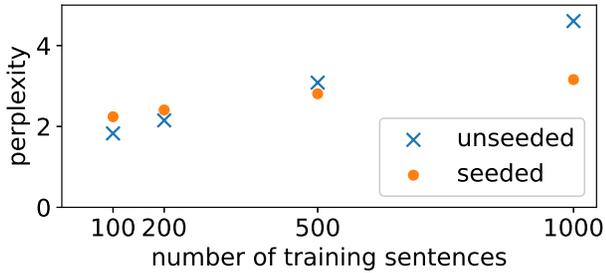}
    \caption[Perplexity with and without seed]{Perplexity vs.\ number of sentences comparing the performance with and without a seed from the synthetic data.}
    \label{fig:seeding}
\end{figure}
Suppose we have a solution with a desirable structure but a limited vocabulary. The synthetic grammar we introduced in section \ref{subsec:synth} is a good example. We can use one of its solutions as a seed for solving the COCA data set. \new{Figure \ref{fig:seeding} shows how the perplexity can grow rapidly without the seed. Of course, this is also without the benefit of the relaxation-based techniques introduced in section \ref{subsec:relax}. A good seed can be an alternative to relaxing constraints. Instead of trying to accommodate the noise, it encourages the algorithm to identify the parts of the data that are consistent with the structure of the seed. The perplexity grows much more slowly when the algorithm is guided by this helping hand.}

\section{Conclusion}

The current state of natural language processing research reflects a disconnect between idealized instructive models that computational linguists use to explain the structure of language and large neural networks that computer scientists use to imitate patterns within large data sets. The former are well-structured and relatively easy to interpret and elucidate a clear connection between form (syntax) and meaning (semantics), but they can be clunky to embed into an architecture that is compact and computationally efficient, especially when the word categories must be discovered without supervision. The latter do remarkably well at producing human-like output, but without transparent machinery and with a troublesome capacity to fabricate things that are syntactically sound but semantically false. The method presented here is an attempt to bridge this disconnect by encoding word categories in a way that preserves the discrete logical structure of a categorial grammar or pregroup grammar and allows efficient computations to discover categories from small data sets and produce realistic novel sentences.

The design of large language models steers them away from simple logical rules in favor of subtle statistical patterns gleaned from their enormous heaps training data. It does not resemble the way humans infer such logical relationships from comparatively minuscule amounts of exposure. We do not claim to fully understand human cognition, but LGE certainly tracks more closely with models constructed by humans to explain the logical structure of language, and this structure allows us to make inferences from as few as a hundred sentences.
\new{The word categories inferred from synthetic data closely align with linguists' expectations, but they can also tell us about symmetries of the syntax. In this capacity LGE might be useful for linguists studying categorial or pregroup grammars. The word categories inferred from natural data are, naturally, a bit messier, but still not too difficult to interpret and can yield insights into which words or constructions are important in parsing the data. 
Surprises can happen with LGE, especially when we use it to generate novel sentences, but tracing their origins is trivially easy thanks to the model's transparency and ability to work with small amounts of data. Using a ``seed'' consisting of a limited lexicon can help as the amount of data increases, giving the algorithm a head start on its parses and encouraging it to add new vocabulary to the existing structure.
The fact that some sentences fail to parse can also be informative, as it indicates something worth looking at in the training data --- a possible grammatical mistake or unusual syntactic construction, for instance. Conversely, we can relax constraints to make the model more accepting of unusual constructions and increase the parse success rate.}



The approach that we have presented here is undoubtedly in its infancy. More work would be needed for it to come anywhere near the sophistication of today's large language models, but it should not be a rivalry. Our hope is that LGE can provide a more rigid, transparent, and logical framework for language modeling. One can incorporate statistical features on top of this framework; indeed, we hinted at this possibility with the simple probabilistic context rules we used for generating sentences, but there is no reason that more refined statistics cannot be involved. Conversely, one can imagine using LGE to assist existing language models by helping them make inferences from smaller, more controllable data sets, or by discovering logical and functional relationships within text that could not only make syntactic parsing more interpretable but also enable semantic parsing to address the hallucination problem. For a model that starts without knowledge of linguistic structure and treats language as a predict-the-next-word problem, the path for improving the model is to increase the number of constraints on its parameters by adding more training data. The path for LGE is in many ways the opposite: The built-in structure already imposes tight constraints to the point that a small amount of data is enough to learn syntax. The challenge will be in finding the best ways to relax constraints and to incorporate this discrete structure into larger, more continuous language models without losing the rigidity that makes it useful. 

\section{Acknowledgements}

The authors wish to thank Jonathan Yedidia for helpful conversations and Morten Christiansen for suggesting the COCA data set.